\pdfoutput=1 
\documentclass{article}

\usepackage[preprint]{neurips_2023}

\usepackage[utf8]{inputenc} 
\usepackage[T1]{fontenc}    
\usepackage{hyperref}       
\usepackage{url}            
\usepackage{booktabs}       
\usepackage{amsfonts}       
\usepackage{nicefrac}       
\usepackage{microtype}      
\usepackage{xcolor}         
\usepackage{graphicx}
\usepackage{epsfig}
\usepackage{amssymb}
\usepackage{bbding}
\usepackage{ragged2e} 
\usepackage{booktabs,makecell, multirow, tabularx}
\usepackage{amsmath}
\usepackage{makecell}
\usepackage{subfigure}
\usepackage{float} 

\title{IDO-VFI: Identifying Dynamics via Optical Flow Guidance for Video Frame Interpolation with Events}

\author{
	Chenyang Shi$^{1,\dag}$\quad Hanxiao Liu$^{1,\dag}$ \quad Jing Jin$^{1, *}$ \quad\textbf{ Wenzhuo Li}$^{1}$\\
	\textbf{Yuzhen Li}$^{1}$ \quad \textbf{Boyi Wei}$^{1}$ \quad \textbf{Yibo Zhang}$^{1}$\\
	$^{\dag}$Equal Contributions $^{*}$Corresponding Author\\
	$^{1}$School of Instrumentation and Opto-electronics Engineering\\
	$^{1}$Beihang University\\
	$^{1}$Beijing, P.R. China, 100191 \\
	\texttt{\{jinjing\}@buaa.edu.cn} \\  
}

\begin{document}
	
	\maketitle

	\begin{abstract}
		Video frame interpolation aims to generate high-quality intermediate frames from boundary frames and increase frame rate. While existing linear, symmetric and nonlinear models are used to bridge the gap from the lack of inter-frame motion, they cannot reconstruct real motions. Event cameras, however, are ideal for capturing inter-frame dynamics with their extremely high temporal resolution. In this paper, we propose an event-and-frame-based video frame interpolation method named IDO-VFI that assigns varying amounts of computation for different sub-regions via optical flow guidance. The proposed method first estimates the optical flow based on frames and events, and then decides whether to further calculate the residual optical flow in those sub-regions via a Gumbel gating module according to the optical flow amplitude. Intermediate frames are eventually generated through a concise Transformer-based fusion network. Our proposed method maintains high-quality performance while reducing computation time and computational effort by 10\% and 17\% respectively on Vimeo90K datasets, compared with a unified process on the whole region. Moreover, our method outperforms state-of-the-art frame-only and frames-plus-events methods on multiple video frame interpolation benchmarks. Codes and models are available at  \url{https://github.com/shicy17/IDO-VFI}.
	\end{abstract}

	\section{Introduction}

	Video frame interpolation (VFI) increases the video frame rate by inserting a reconstruction frame into two consecutive frames. Due to the limitation of the fixed frame rate of ordinary camera, the frame-only video frame interpolation methods inevitably lose the dynamics in the interval between consecutive frames. In order to compensate for the lack of inter-frame information, motion models are often used, but those models cannot account for the real motions. 
	
	Event cameras \cite{posch2014retinomorphic} are bio-inspired vision sensor, each pixel of which independently perceives and encodes relative changes in light intensity. Event cameras output sparse, asynchronous streams of events instead of frames, with advantages of high temporal resolution, high dynamics, and low power consumption. An event is usually expressed as a tuple $e=(x,y,p,t)$, which means that at timestamp $t$, an event with polarity $p\in\{-1, 1\}$ is generated at the pixel $(x,y)$. Positive polarity indicates that the change of light intensity from week to strong is beyond the threshold, while negative polarity is just the opposite. Because an event camera has high temporal resolution up to microseconds, it can capture complete changes or motion between frames. 
	
	The event flow is the embodiment of inter-frame changes. Therefore, the optical flow estimated from the events does not require any motion model to be fitted, which can be inherently nonlinear. Since events lack intensity information, frame-based optical flow is complementary to event-based optical flow. By combining these two kinds of optical flow, more accurate estimation results can be obtained. Meanwhile, it is possible to reconstruct high-quality keyframes at any timestamp, since real inter-frame dynamics are captured.
	
	\begin{figure}[t]
		\begin{center}
			\subfigure[PSNR/Runtime/Parameters comparison]{
				\includegraphics[width=0.51\linewidth]{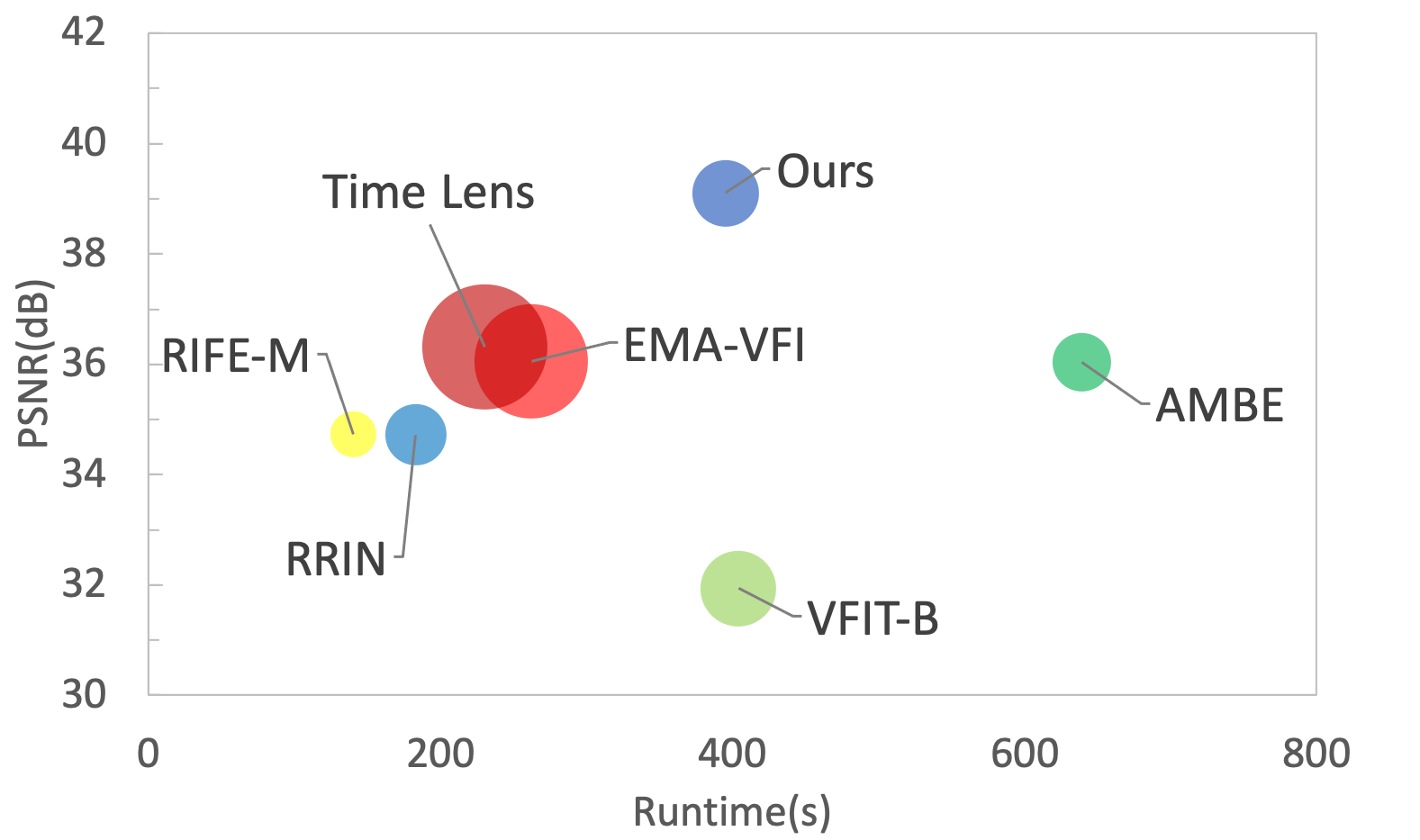}
			\label{01a}}
			\subfigure[Results of proposed each module on Vimeo90K]{
				\includegraphics[width=0.45\linewidth]{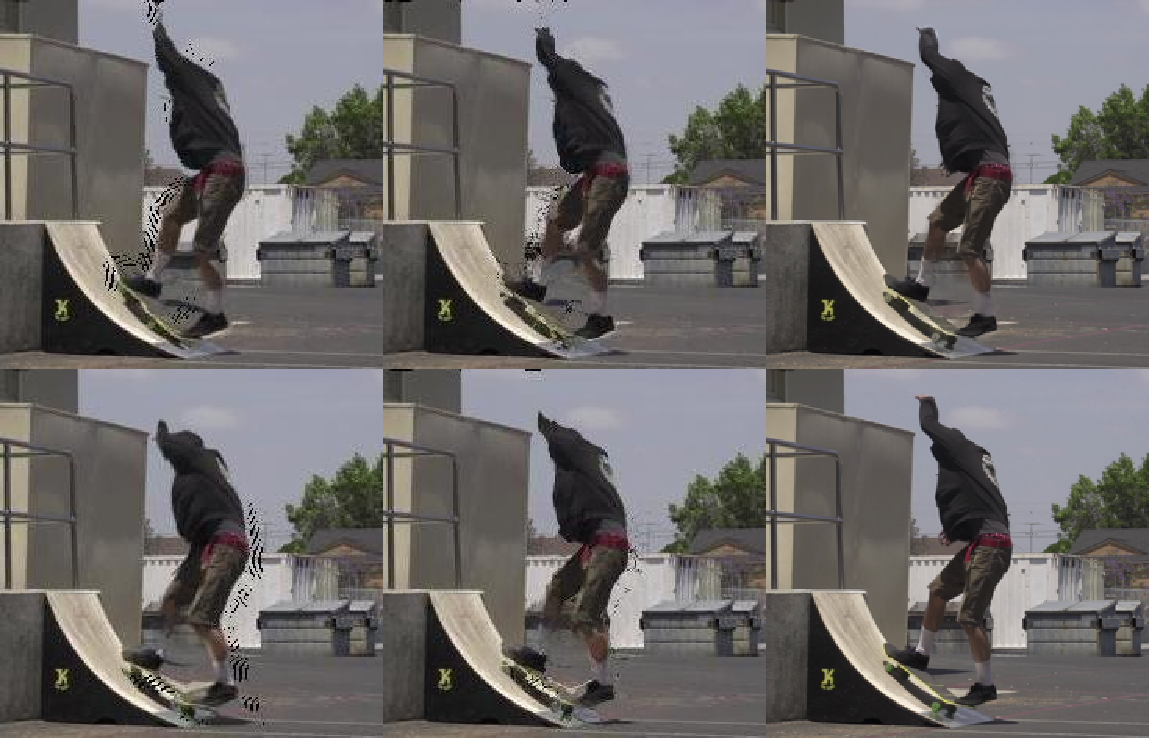}
			\label{01b}}
		\end{center}
		\caption{(a) The quantitative PSNR/Runtime/Parameters comparison of ours and state-of-the-art VFI methods on Vimeo90K. The larger circle represents larger number of parameters. Our proposed method achieves the best performance with relatively small model parameters and fast computation time. (b) Results of proposed each module on Vimeo90K. The first column shows the warping-based results of optical flow estimation module. The second column presents refined frames generated from residual flow estimation module. Right top is the final output, and right bottom is the ground-truth. The quality of output improves step by step.}
		\label{01}
	\end{figure}
	
	Furthermore, the real inter-frame motion information lays the foundation for reliable differential processing of different image regions. For pixel areas with small motion amplitudes, only a simple estimation is needed to obtain an accurate optical flow field. However, for complex dynamic regions where
	simple optical flow estimation is insufficient to address the problem, further estimation of residual optical flow is required in order to obtain more accurate results. With the help of event flow, we can easily distinguish dynamic and static areas in the image, and adopt different optical flow estimation strategies, which can greatly reduce the amount of calculation while maintaining high-precision results. Our main contributions are as follows:
	\begin{itemize}
		\item  A novel and trainable optical flow guidance mechanism for identifying the dynamics of the boundary frames and events is proposed, considering the corresponding relationship between adjacent dynamic regions.
		\item  We propose an event-based residual optical flow estimation method to further dynamically evaluate the optical flow field, of which the computation time and computational effort are reduced by 10\% and 17\% respectively, while the performance is almost the same as processing the whole image without distinction.
		\item Our proposed method achieves state-of-the-art results on multiple benchmark datasets compared to frame-only and events-plus-frames VFI methods. Codes and models are available publicly.
	\end{itemize}
	This paper is organized as follows. First, the main work in this field is introduced. Second, our proposed VFI method and its components are present. Third, the experimental details and quantitative results are illustrated. Subsequently, ablation experiments are conducted. Finally, the paper is summarized and discussed.
		\begin{figure}[ht]
		\subfigure[Ground Truth]{
			\begin{minipage}[htbp]{0.182\linewidth}
				\centering
				\includegraphics[height=1.58cm]{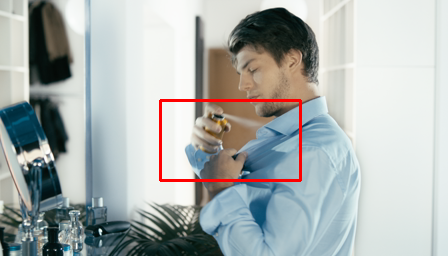}\vspace{0.18pt}
				\includegraphics[height=1.58cm]{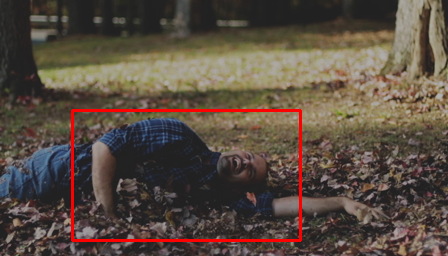}\vspace{0.18pt}
				\includegraphics[height=1.58cm]{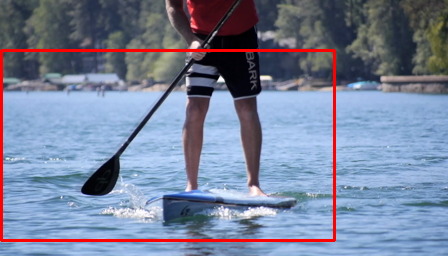}\vspace{0.18pt}
				\includegraphics[height=1.58cm]{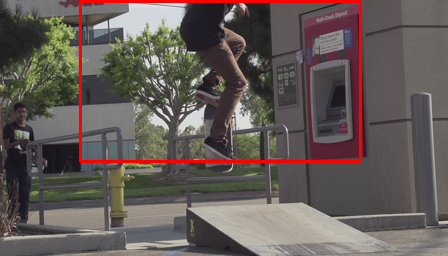}\vspace{0.18pt}
				\includegraphics[height=1.58cm]{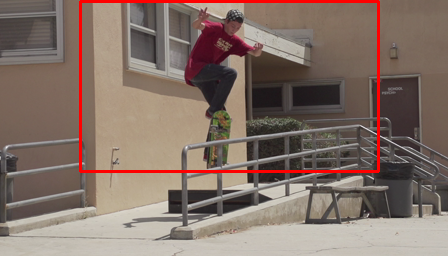}\vspace{0.18pt}
			\end{minipage}
		}
		\subfigure[RIFE]{
			\begin{minipage}[htbp]{0.182\linewidth}
				\centering
				\includegraphics[height=1.58cm]{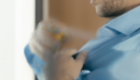}\vspace{0.18pt}
				\includegraphics[height=1.58cm]{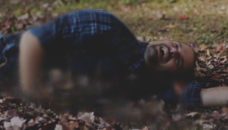}\vspace{0.18pt}
				\includegraphics[height=1.58cm]{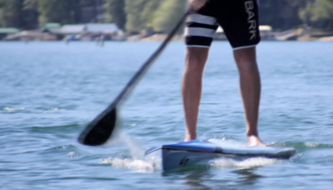}\vspace{0.18pt}
				\includegraphics[height=1.58cm]{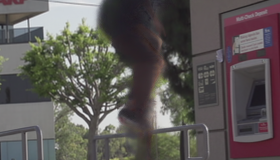}\vspace{0.18pt}
				\includegraphics[height=1.58cm]{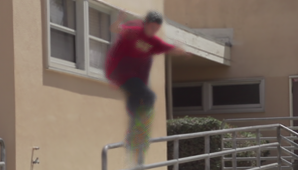}\vspace{0.18pt}
			\end{minipage}
		}
		\subfigure[AMBE]{
			\begin{minipage}[htbp]{0.182\linewidth}
				\centering
				\includegraphics[height=1.58cm]{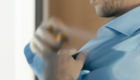}\vspace{0.18pt}
				\includegraphics[height=1.58cm]{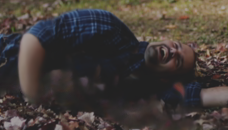}\vspace{0.18pt}
				\includegraphics[height=1.58cm]{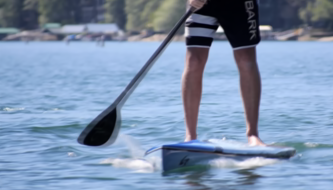}\vspace{0.18pt}
				\includegraphics[height=1.58cm]{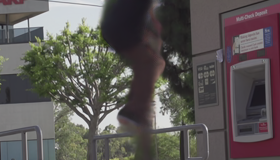}\vspace{0.18pt}
				\includegraphics[height=1.58cm]{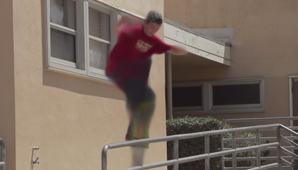}\vspace{0.18pt}
			\end{minipage}
		}
		\subfigure[EMA-VFI]{
			\begin{minipage}[htbp]{0.182\linewidth}
				\centering
				\includegraphics[height=1.58cm]{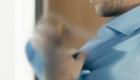}\vspace{0.18pt}
				\includegraphics[height=1.58cm]{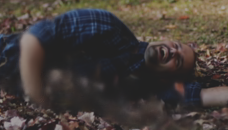}\vspace{0.18pt}
				\includegraphics[height=1.58cm]{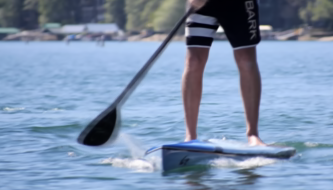}\vspace{0.18pt}
				\includegraphics[height=1.58cm]{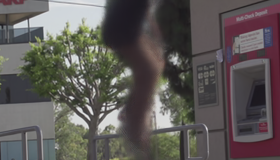}\vspace{0.18pt}
				\includegraphics[height=1.58cm]{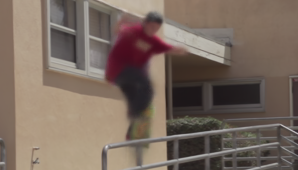}\vspace{0.18pt}
			\end{minipage}
		}
		\subfigure[Ours]{
			\begin{minipage}[htbp]{0.182\linewidth}
				\centering
				\includegraphics[height=1.58cm]{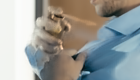}\vspace{0.18pt}
				\includegraphics[height=1.58cm]{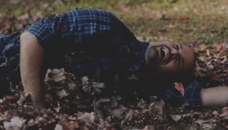}\vspace{0.18pt}
				\includegraphics[height=1.58cm]{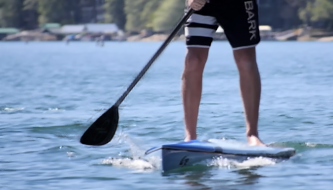}\vspace{0.18pt}
				\includegraphics[height=1.58cm]{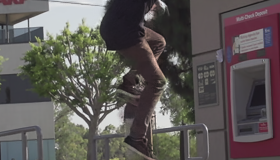}\vspace{0.18pt}
				\includegraphics[height=1.58cm]{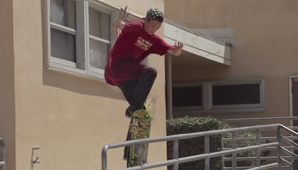}\vspace{0.18pt}
			\end{minipage}
		}
		\caption{Visual comparison among our proposed method, the state-of-the-art methods RIFE, AMBE, EMA-VFI on the Vimeo90K-triplet validation set. }
		\label{c}
	\vspace{-0.8cm}
	\end{figure}

	\section{Related Works}
	Reconstructing dynamics and luminosity is the key task of VFI. Thus, the warping-based methods and synthesis-based methods become the mainstream methods of VFI.
	
	\textbf{Frame-only VFI Methods.} Warping-based methods use photometric consistency assumptions to estimate inter-frame motion, which is very effective for video sequences with short inter-frame blind times and simple motion, but it only warps pixels and cannot reconstruct photometric information.
	The original methods usually assume that the optical flow between frames is first-order, such as \cite{sun2018pwc, jiang2018super, bao2019depth, park2020bmbc, niklaus2020softmax, sim2021xvfi, yu2022deep}.Meanwhile, several complex motion models have been proposed. Xu et al.\cite{xu2019quadratic} proposed a method for estimating the secondary optical flow, but this method needs to input four key frames at a time. Park et al. proposed AMBE \cite{park2021asymmetric}, on the basis of BMBC \cite{park2020bmbc}, using anchor frames to estimate asymmetric motion without relying on linear optical flow assumptions. However, the assumed motion models may fail once the actual motion becomes complex.
	
	Synthesis-based methods \cite{lee2020adacof, choi2021motion, shi2022video, niklaus2023splatting} directly fuses the image features of boundary frames to generate intermediate frames, which can reconstruct photometric information. However, the synthesis method performs poorly when there is complex motion in the time interval. In order to restore this defect, it usually takes multiple consecutive frames, e.g. four frames \cite{shi2022video}, as input. Some models \cite{shangguan2022learning, oh2022demfi} combine warping-based and synthesis-based methods, considering complementarity between the two, which can reconstruct dynamics and photometry while estimating inter-frame motion.
	 
	\textbf{Events-plus-frames VFI Methods.} In recent years, there have been attempts \cite{pan2019bringing, wang2020joint, yu2021training, tulyakov2021time, tulyakov2022time, he2022timereplayer, zhang2022unifying, wu2022video} to combine frames and events for VFI. Event-based optical flow can still be accurately estimated under condition of complex intermediate motions, since event-based optical flow estimation is not based on linearity assumptions. Although event cameras do not encode photometric information, they are complementary to frame-based cameras. Yu et al. \cite{yu2021training} respectively extracted the multi-scale features of events and frames for fusion, and proposed a sub-pixel-level attention mechanism, which uses event information to supplement inter-frame information to achieve weakly supervised learning. Tulyakov et al. \cite{tulyakov2021time} proposed Time Lens, which combines events and frames to generate warping-based and synthesis-based images respectively, and outputs the final result through an attention-based network. However, it has a very large number of model parameters. On the ground of \cite{tulyakov2021time}, Tulyakov et al. proposed Time Lens++ \cite{tulyakov2022time}, which encodes optical flow as cubic splines and warps the features for fusion in an encoder-decoder network. Unfortunately, the amount of model parameters is still large. He et al. proposed TimeReplayer \cite{he2022timereplayer}, an event-based unsupervised video frame interpolation method. The unsupervised learning method decreases the dependency on the use of high frame-rate datasets. Although these methods achieve good performance, they are computationally expensive and do not maximize the advantages of the properties of events to characterize motion.
	
	Combining events and frames can estimate the complete inter-frame motion without any motion model, so the motion amplitude of all pixel regions can be obtained. Simple processing is enough for areas with small motion amplitude. Only areas with large motion amplitude require more complex processing. Therefore adopting different calculation strategies for pixel regions with different motion amplitude can save calculation, while maintaining the quality of the output.  Some frame-only VFI methods for reducing the computational overhead have been proposed. Choi et al. \cite{choi2021motion} proposed a method to evaluate the motion of the local area, reasonably select the model depth to process the local area, or perform downscale processing on the local area at different scales, so as to reduce the computational overhead. But this method is only based on the assumption of photometric consistency and cannot cover complex motions. Therefore, we propose an event-based VFI method that reduces computation time and overhead by dynamically estimating residual optical flow in pixel regions while maintaining high-quality output.
	
	\section{Proposed Method}
	\subsection{Problem Formulation}
	Assuming that we are given two consecutive frames $I_0$ and $I_1$ at time 0 and 1, as well as events sequences $E_{0\rightarrow1}$ consisting all events triggered between the interval. The task is generating intermediate frame $\hat{I_{t}}$ at arbitrary time $t$, where $t\in [0,1]$. Besides, according to the interpolating timestamp $t$, we can divide the event sequences $E_{0 \rightarrow 1}$ into two parts $E_{0 \rightarrow t}$ and $E_{t \rightarrow 1}$. The event sequence $E_{a \rightarrow b}$ is represented as a voxel grid $V_{a \rightarrow b}$ \cite{zihao2018unsupervised}.
	\subsection{Overview}
	The proposed framework is mainly consisted of four components: Optical flow estimation module, Gumbel gating module, Residual optical flow estimation module and Transformer-based fusion module. 
	
\begin{figure}[htbp]
		\begin{center}
			\includegraphics[width=1\linewidth]{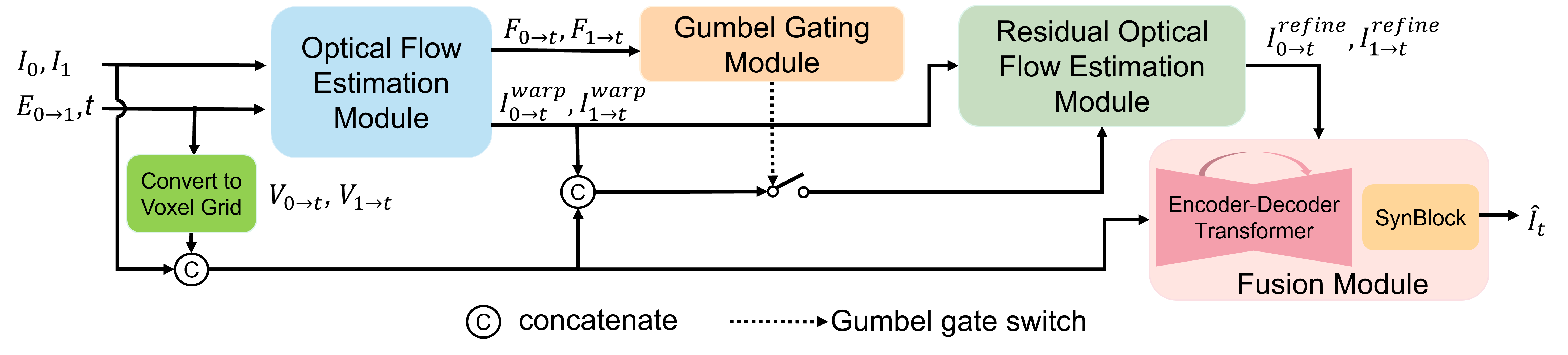}
		\end{center}
		\caption{Overview of proposed model architecture.}
		\label{1}
	\end{figure}

	First, the existing consecutive frames $I_0$, $I_1$ and event sequence are input to the optical flow estimation network for calculating bidirectional optical flows $F_{0 \rightarrow t}$ and $F_{1 \rightarrow t}$. The $I_0$ and $I_1$ are warped according to these optical flows to generate intermediate frames $I^{warp}_{0}$ and $I^{warp}_{1}$. Subsequently, the boundary frames are evenly divided into multiple sub-regions, which are further categorized as dynamic or static regions according to the Gumbel gating network. The dynamic regions in the image will be fed into the residual optical flow estimation network to estimate the residual optical flows. Then warping the $I^{warp}_{0}$ and $I^{warp}_{1}$ to generate $I_{0 \rightarrow t }^{refine}$ and $ I_{1 \rightarrow t }^{refine}$. The final output of the proposed model is generated by synthesizing the boundary frames and warping-based frames. Unnecessary computing costs of static region could be significantly reduced through this method, while maintaining high-quality final result of the network. The overall architecture is illustrated in Figure.\ref{1}. 
	\subsection{Optical Flow Estimation}
	A UNet \cite{jiang2018super} is adopted as the backbone of optical flow estimation network, and extended by us for event sequence inputting. Note that the network performs symmetric processing for calculating $F_{0 \rightarrow t}$ and $F_{1 \rightarrow t}$,  we thus only introduce the processing for calculating $F_{0 \rightarrow t}$. The flow network extracts feature representations from both input frames $I_0$, $I_1$ and events $V_{0 \rightarrow 1}$. In addition, inspired by Time Lens++\cite{tulyakov2022time}, we compute cubic motion splines $\{S_{0 \rightarrow 1}^{\Delta x},S_{0 \rightarrow 1}^{\Delta y}\}$ for each location instead of linear optical flow. These cubic splines which are presented by \textit{K}-th control points in order to model horizontal and vertical displacement of each pixel of previous frame as a function of time. 	
	\begin{figure}[tbp]
		\begin{center}
			\subfigure[Region division]{
				\includegraphics[height=5cm]{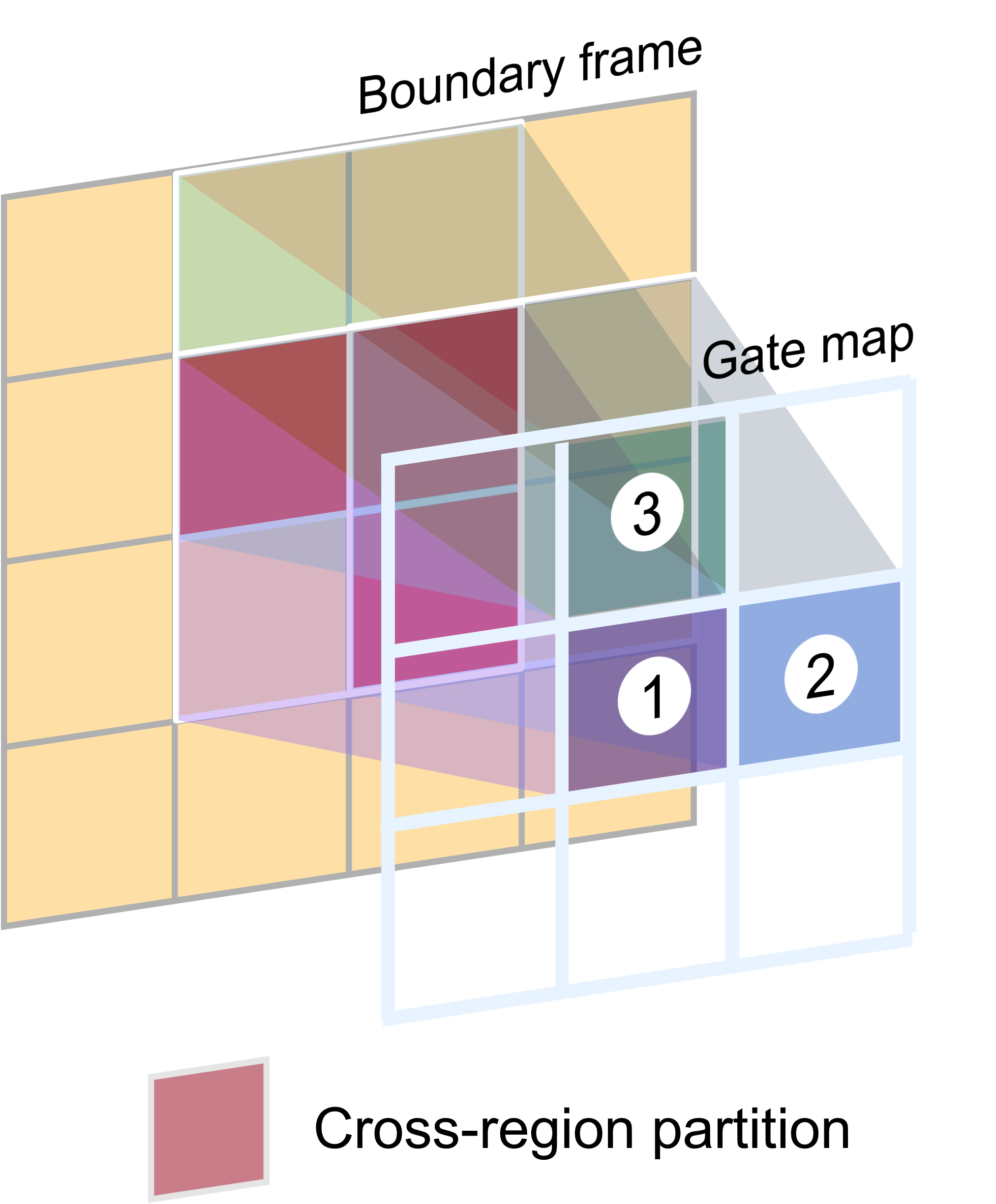}
				\label{4a}}
			\subfigure[Gating network]{
				\includegraphics[height=5cm]{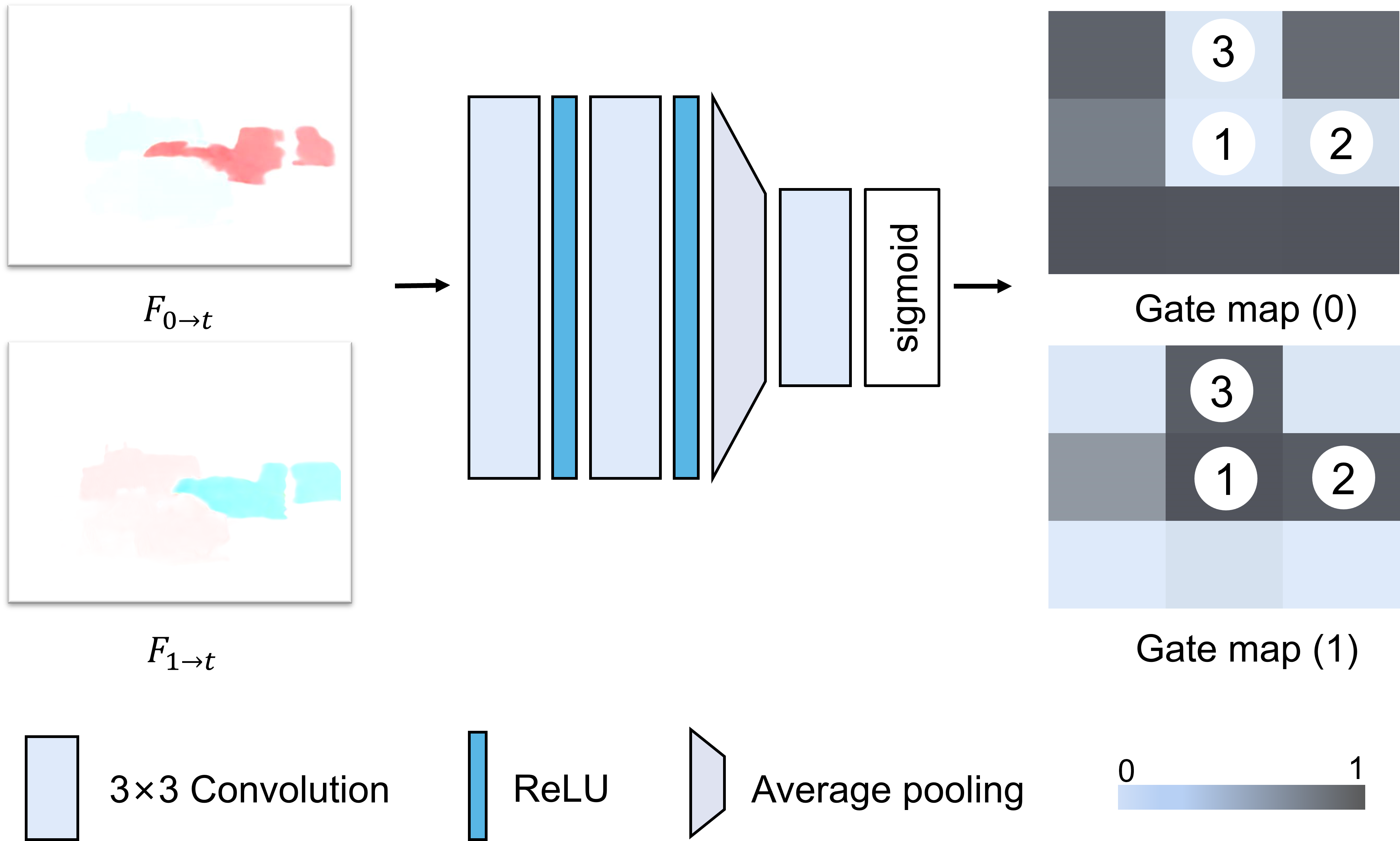}
				\label{4b}}
		\end{center}
		\caption{The proposed Gumbel gating module.}
		\label{4}
	\end{figure}Optical flows could be obtained by sampling from the motion splines, which reduce the computation cost from $\mathcal{O}(N)$ to $\mathcal{O}(1)$ for the calculation of optical flows. By adding the information of events from blind time, the real motion can be modeled in the flow network. As a result, a nonlinear optical flow $F_{0 \rightarrow t}$ for random time $t$ is obtained by sampling from the motion spline $F_{0 \rightarrow 1}$ with minimal additional computational cost. 
	
	Meanwhile, the intermediate frames $I^{warp}$ are obtained by warping the boundary frames using the estimated optical flow, and described as
	\begin{equation}
		I_0^{warp}=W_f(I_0, F_{0 \rightarrow t})
	\end{equation}
	\begin{equation}
		I_1^{warp}=W_f(I_1, F_{1 \rightarrow t})
	\end{equation}
	where $W_f(\cdot)$ is the softmax-splatting forward warping operation \cite{niklaus2020softmax}. Note that, since the estimated optical flow is forward, this forward warping operation is employed.

	\subsection{Gumbel Gating Module}
	On the basis of roughly estimating the bilateral optical flow $F_{0 \rightarrow t}$ and $F_{1 \rightarrow t}$, we then divide the dynamic and static regions in boundary frames. The discrimination of the region type is performed by calculating a Bernoulli probability distribution generated from a trainable Gumbel gating mechanism. 
	
	Pixel regions with high-magnitude optical flow field or violent motion will be considered as dynamic regions, and conversely, regions with slight optical flow changes or smooth motion will be considered as static regions. If the whole image is divided into dynamic and static regions according to the optical flow of each pixel, a large number of discrete and irregularly shaped pixel blocks will inevitably appear, which is difficult for further processing. In order to simplify the process, we first set an adjustable rectangular sliding window whose length and width are set to W/2 and H/2 of the input optical flow field respectively. This box will start scanning from the upper left corner of the input, and the horizontal and vertical stride are W/4 and H/4, individually. 
	
	As a result, the sliding window operation generates a total of nine pixel regions $R^{0,1}_{i},i=0,1,\cdots,8$ for each boundary frame, and adjacent pixel regions have a cross-region partition, as shown in Figure.\ref{4a}. The number of these pixel areas is adjustable. Note that we differ from Choi et al. \cite{choi2021motion} in that we take into account the connections between adjacent pixel regions. It divides the image into several regions evenly, and the network determines the number of layers that each region needs to process. In contrast, our proposed method preserves the correlation between sub-graphs. 
	
	Subsequently,  optical flow $F_{0 \rightarrow t}$ and $F_{1 \rightarrow t}$ are input into a lightweight gating network to generate a gate map $M_{2\times3\times 3}$. Each pixel on the gate map is a Bernoulli distribution, representing the probability that the corresponding sub-region belongs to the dynamic region or the static region. We can get a binary mask $P$ by rounding the gate map.
	\begin{equation}
		P = G(M_{2\times3 \times 3})
	\end{equation}
	Where G($\cdot$) is rounding operation, but when training, G($\cdot$) is Gumbel-softmax operation \cite{jang2017categorical}. The Gumbel-softmax tricks solve the problem that binarization is not differentiable. Note that the final decision is based on the binary mask $P$. The structure of the gating network is shown in Figure.\ref{4b}.

	\subsection{Residual Optical Flow Estimation}
	All the dynamic regions judged by the Gumbel gating network will be fed into the residual optical flow estimation module to further estimate the residual optical flow. 
	
	For a dynamic region $R_{i}^{single},i= 0,1,2,\cdots,n $, its corresponding optical flow field $F_{i}^{single}$ and the corresponding parts of $I_{0}, I_{1}, V_{0 \rightarrow t}, V_{1 \rightarrow t}, I_0^{warp}$ and $I_1^{warp}$ are first concatenated and input into the residual optical flow estimation network. Subsequently, the network will output sub-region refined optical flows $F_{i,1 \rightarrow t}^{refine}$ and $F_{i,0 \rightarrow t}^{refine}$. 	Secondly, the refined optical flow of each dynamic region will be padded with 0 to the same size as the boundary frame. They will be fed into an attention-based network to generate a weight map, which is used to calculate the optical flow field coefficient of the corresponding part in each dynamic region that produces the cross-region partition. Thirdly, the residual optical flows in these partitions will be montaged with that of other dynamic regions, and the entire refined optical flows $F^{refine}_{0 \rightarrow t}$ and $F^{refine}_{1 \rightarrow t}$ will be output. Finally, $I_0^{warp}$ and $I_1^{warp}$ are processed by backward warping $W_b(\cdot)$, and output $I_{0 \rightarrow t }^{refine}$ and $I_{1 \rightarrow t}^{refine}$, described as follows.
	\begin{equation}
		I_{0 \rightarrow t }^{refine}=W_b(I_0^{warp}, F_{0 \rightarrow t})
	\end{equation}
	\begin{equation}
		I_{1 \rightarrow t }^{refine}=W_b(I_1^{warp}, F_{1 \rightarrow t})
	\end{equation}
	
\begin{figure}[htbp]
		\begin{center}
			\includegraphics[width=1\linewidth]{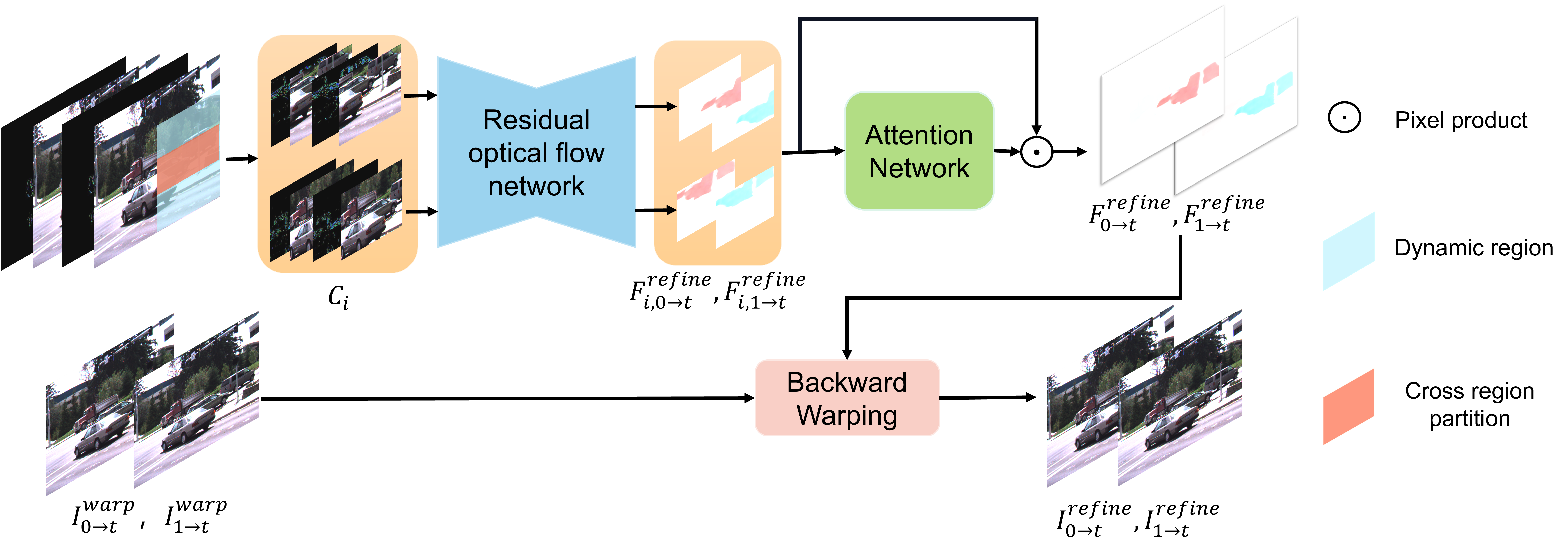}
		\end{center}
		\caption{The proposed residual optical flow estimation network.}
		\label{5}
	\end{figure}

	Note that backward warping is used to save computation time. Both the residual optical flow estimation network and the attention network are constructed by a UNet, and the entire architecture of residual optical flow estimation module is shown in Figure.\ref{5}.

	\subsection{Transformer-Based Fusion}
	The boundary frames $I_0$, $I_1$, event streams $E_{0 \rightarrow t}$, $E_{1 \rightarrow t}$, and the warping-based frames $I_{0 \rightarrow t }^{refine}$, $ I_{1 \rightarrow t }^{refine}$ are fed into a Transformer-based encoder-decoder network. This synthesis-based network is modified from VFIT-B\cite{shi2022video}. Note that the differences between our method and VFIT-B. First, we only use one SynBlock for fusion. While VFIT-B applies three SynBlocks to achieve feature fusion on three scales. Furthermore, the input for VFIT-B are four consecutive keyframes $I_0$, $I_1$, $I_2$, $I_3$. Moreover, VFIT-B only generates one intermediate frame at $t_{0.5}$, while our proposed method can generate intermediate frames at any timestamp. Fortunately, with the help of $I_{0 \rightarrow t }^{refine}$, $ I_{1 \rightarrow t }^{refine}$, we generate results that outperform other methods with a concise fusion network. The output of the block is the final result $I_{t}$, and described as follows.
	 	\begin{equation}
	 	I_{t} = S(I_0, I_1, E_{0 \rightarrow t}, E_{1 \rightarrow t}, I_{0 \rightarrow t }^{refine}, I_{1 \rightarrow t }^{refine})
	 \end{equation}
	where S($\cdot$) is fusion operation.
	\begin{table*}[tbp]
		\scriptsize	\caption{Quantitative comparison of the proposed method with the state-of-the-art methods on multiple VFI benchmarks. Vimeo-3f and Vimeo-7f indicate Vimeo90K-Triplet-Train dataset and Vimeo90K-Septuplet-Train dataset respectively. The best PSNR(dB)\textbackslash SSIM result is \textcolor{red}{boldfaced}, and the second best is  \textcolor{blue}{\underline{underlined}}. $\ddagger$ indicates that we quote the results in corresponding papers. 
		}
		\label{tab4}
		\centering
		\setlength{\tabcolsep}{1.6pt}
		\renewcommand{\arraystretch}{1.2} 
		\begin{tabular}{lccccccccccccccc}
			\hline
			\toprule
			\multirow{3}*{Method}  &\multirow{3}*{Year}   &\multirow{3}*{\makecell{Training \\ dataset}} &\multirow{3}*{Frame} &\multirow{3}*{Event} &\multicolumn{2}{c}{\makecell{Vimeo90K-\\Triplet-Test \cite{xue2019video}}} &\multicolumn{4}{c}{Middleburry \cite{baker2011database}} &\multicolumn{4}{c}{GoPro\cite{nah2017deep}} &\multirow{3}*{\makecell{\#Param. \\ (Million)}}  \\
			\cline{6-15}
			\multicolumn{5}{c}{} &\multicolumn{2}{c}{1 frame skip}  &\multicolumn{2}{c}{1 frame skip} &\multicolumn{2}{c}{3 frames skip} &\multicolumn{2}{c}{7 frames skip} &\multicolumn{2}{c}{15 frames skip}\\
			\cline{6-15}
			\multicolumn{5}{c}{} & PSNR & SSIM & PSNR & SSIM &PSNR &SSIM& PSNR & SSIM &PSNR&SSIM\\
			\midrule
			E2VID \cite{rebecq2019events} &2019&MS-COCO &\XSolidBrush &\Checkmark &10.77&0.361&11.51&0.377&11.39&0.375&11.8&0.482&12.11&0.472  &10.71\\
			DAIN \cite{bao2019depth}  &2019&Vimeo-3f&\Checkmark &\XSolidBrush&34.20  &0.962   &30.87  &0.899 &26.67 &0.838& 28.81 &0.876 &24.39 &0.736  &24.03 \\
			RRIN \cite{li2020video} &2020&Vimeo-3f&\Checkmark &\XSolidBrush&34.72  &0.962  &31.08&0.896 &27.18  &0.837&28.96 &0.876 &24.32  &0.749 &19.19  \\
			BMBC \cite{park2020bmbc} &2020&Vimeo-3f &\Checkmark &\XSolidBrush &34.56  &0.962  &30.67 &0.885  &26.86  &0.834&29.08 &0.875  &23.68  &0.736 &11.00 \\
			AMBE \cite{park2021asymmetric} &2021&Vimeo-3f &\Checkmark &\XSolidBrush&36.04 &\textcolor{blue}{\underline{0.969}}  &31.72 &0.908&26.64&0.833&30.84 &0.925&26.12&0.857&18.10\\
			VFIT-B \cite{shi2022video} &2022&Vimeo-7f&\Checkmark &\XSolidBrush&31.94 &0.926  &28.37 &0.863 &-  &-  &-&-&-&- &29.00\\
			RIFE \cite{huang2022real}&2022&Vimeo-3f&\Checkmark &\XSolidBrush &34.73&0.960&31.40&0.901&27.97&0.849&32.23&0.937&28.82&0.892&10.71\\
			EMA-VFI\cite{zhang2023extracting}&2023&Vimeo-3f&\Checkmark &\XSolidBrush &36.05&0.968&32.06&0.909&28.67&0.860&32.79&0.942&29.70&0.904&65.66\\
			Time Lens \cite{tulyakov2021time} $\ddagger$&2021&Vimeo-3f &\Checkmark &\Checkmark&\textcolor{blue}{\underline{36.31}}  &0.962 &\textcolor{blue}{\underline{33.27}} &\textcolor{blue}{\underline{0.929}} &\textcolor{blue}{\underline{32.13}}&\textcolor{blue}{\underline{0.908}} &\textcolor{blue}{\underline{34.81}}&0.959&\textcolor{blue}{\underline{33.21}}&\textcolor{blue}{\underline{0.942}} &79.20\\
			TimeReplayer \cite{ he2022timereplayer}$\ddagger$&2022&Vimeo-3f &\Checkmark &\Checkmark&35.12&0.963 &32.74&0.912&30.91 &0.887&34.02&\textcolor{blue}{\underline{0.960}}&-&-&-\\
			\textbf{Ours}    &2023&Vimeo-3f&\Checkmark &\Checkmark &\textcolor{red}{\textbf{39.10}}&\textcolor{red}{\textbf{0.976}}&\textcolor{red}{\textbf{34.96}}&\textcolor{red}{\textbf{0.948}}&\textcolor{red}{\textbf{32.19}}&\textcolor{red}{\textbf{0.927}}&\textcolor{red}{\textbf{36.04}}&\textcolor{red}{\textbf{0.962}}&\textcolor{red}{\textbf{33.27}}&\textcolor{red}{\textbf{0.944}}&22.63\\
			\bottomrule
			\hline
		\end{tabular}
	\end{table*}
	
	\section{Experiments}
	In this section, the implementation details of the proposed method are first described. Subsequently, the datasets for validation are introduced. Next, the comparison results of the proposed method with other state-of-the-art VFI methods are presented. Finally, ablation studies are conducted to demonstrate the effect of each part of the proposed method.

	\subsection{Implementation Details}

\textbf{Loss Function.} The loss function $\mathcal{L} $ is set as a superposition of L1 loss $\mathcal{L}_1$ and FLOPs $\mathcal{G}$. $\mathcal{L} $ is described as $\mathcal{L} = \mathcal{L}_1+\lambda\mathcal{G}$ \cite{choi2021motion},  where $\lambda$ is a hyper-parameter. In our experiments, $\lambda$ is set as 2e-4, which is a trade-off between computation efficiency and performance.
	
\textbf{Training Method.} The proposed method is trained on the Vimeo90K-Triplet-Train dataset \cite{xue2019video}, following the popular paradigm that other VFI approaches \cite{park2020bmbc, park2021asymmetric, tulyakov2021time, huang2022real, park2023biformer,zhang2023extracting} adopted. Because the frame-only datasets do not contain events, we employ the ESIM simulator \cite{rebecq2018esim} to generate synthetic events. Adam optimizer \cite{kingma2014adam} is used to optimize the network with initial learning rate of 1e-4, which is decreased to 1e-5 after the tenth epoch. Each sub-module is trained for 15 epochs, with a batch size of 4, on Vimeo90k-triplet dataset. Each sub-module is trained individually in sequence, with parameters frozen after training is complete. All training are performed on two NVIDIA Tesla A100 GPUs.
	

	\subsection{Comparisons with State-of-the-art Methods}
	\textbf{Datasets.} 	Frame-only VFI benchmark datasets Vimeo90k \footnote{The license is \url{https://toflow.csail.mit.edu}.} \cite{xue2019video}, Middlebury \footnote{The license is \url{http://vision.middlebury. edu/flflow}.} \cite{baker2011database}, GoPro \footnote{The license is \url{https://github.com/SeungjunNah/DeepDeblur_ release}.} \cite{nah2017deep} and frames-plus-events datasets HighQualityFrames\footnote{The license is \url{https://timostoff.github.io/20ecnn}.} \cite{stoffregen2020reducing}, HS-ERGB \footnote{The license is \url{https://rpg.ifi.uzh.ch/timelens}.} \cite{tulyakov2021time} are selected to validate the performance of ours and state-of-the-art VFI methods.	
	
 The methods that achieve the state-of-the-art results are selected as baselines to compare with the proposed methods, including events-only method E2VID \cite{rebecq2019events}, frames-only methods DAIN \cite{bao2019depth}, RRIN \cite{li2020video}, BMBC \cite{park2020bmbc},  AMBE \cite{park2021asymmetric}, VFIT-B \cite{shi2022video}, RIFE \cite{huang2022real}, EMA-VFI \cite{zhang2023extracting} and frames-plus-events methods Time Lens \cite{tulyakov2021time},  TimeReplayer \cite{ he2022timereplayer}. For evaluation, structural similarity (SSIM) \cite{wang2004image} and peak-signal-to-noise-ratio (PSNR) are used to measure the interpolation quality of our and benchmark methods. Note that SSIM is evaluated by \textit{compare\_ssim} in scikit-image library.
	
	We test the interpolation results of ours and other state-of-the-art methods on Vimeo90K-Triplet validation set for skip1, Middleburry validation set for skip 1 and skip 3, and GoPro validation set for skip 7 and skip 15. The comparison results validated on frame-only datasets are shown in Table.\ref{tab4}. Our proposed method outperforms other benchmark methods on these three datasets. Among them, on the Vimeo90k dataset, PSNR and SSIM of ours are respectively 2.79dB and 0.007 higher than the second place. Moreover, SSIM of ours is 0.019 higher than the second place on the Middleburry dataset. Furthermore, our proposed method has less model parameters while ensuring high-quality performance. The visual comparisons of PSNR/Runtime/Parameters on Vimeo90K are shown in Figure.\ref{01a}.
	
\begin{figure}[htbp]
		\subfigure[Ground Truth]{
			\begin{minipage}[htbp]{0.16\linewidth}
				\centering
				\includegraphics[height=1.78cm]{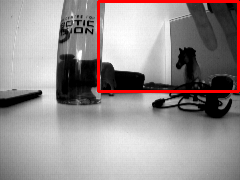}\vspace{0.18pt}
				\includegraphics[height=1.78cm]{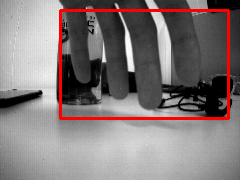}\vspace{0.18pt}
				\includegraphics[height=1.78cm]{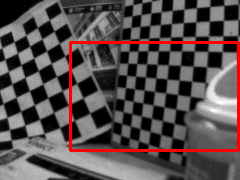}\vspace{0.18pt}
			\end{minipage}
		}
		\subfigure[RIFE]{
			\begin{minipage}[htbp]{0.188\linewidth}
				\centering
				\includegraphics[height=1.78cm]{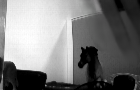}\vspace{0.18pt}
				\includegraphics[height=1.78cm]{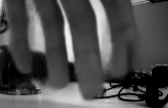}\vspace{0.18pt}
				\includegraphics[height=1.78cm]{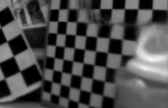}\vspace{0.18pt}
			\end{minipage}
		}
		\subfigure[EMA-VFI]{
			\begin{minipage}[htbp]{0.188\linewidth}
				\centering
				\includegraphics[height=1.78cm]{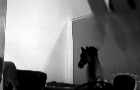}\vspace{0.18pt}
				\includegraphics[height=1.78cm]{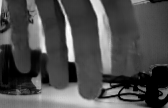}\vspace{0.18pt}
				\includegraphics[height=1.78cm]{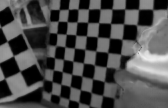}\vspace{0.18pt}
			\end{minipage}
		}
		\subfigure[Time Lens]{
			\begin{minipage}[htbp]{0.188\linewidth}
				\centering
				\includegraphics[height=1.78cm]{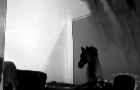}\vspace{0.18pt}
				\includegraphics[height=1.78cm]{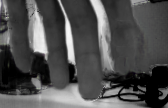}\vspace{0.18pt}
				\includegraphics[height=1.78cm]{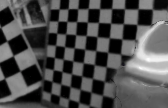}\vspace{0.18pt}
				
			\end{minipage}
		}
		\subfigure[Ours]{
			\begin{minipage}[htbp]{0.188\linewidth}
				\centering
				\includegraphics[height=1.78cm]{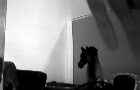}\vspace{0.18pt}
				\includegraphics[height=1.78cm]{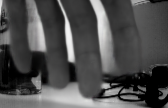}\vspace{0.18pt}
				\includegraphics[height=1.78cm]{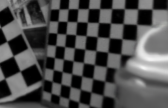}\vspace{0.18pt}
			\end{minipage}
		}
		\caption{Visual comparison among our proposed method, the state-of-the-art methods RIFE, EMA-VFI and Time Lens on HQF dataset containing real-world events.}
		\label{f6}
	\end{figure}

	Due to the discrepancies between real-world and synthetic events, we fine-tune the model trained on synthetic events on datasets containing real-world events. The HighQualityFrames (HQF) \cite{stoffregen2020reducing} and HS-ERGB \cite{tulyakov2021time} datasets are shuffled for fine-tuning, following the paradigm adopted in \cite{tulyakov2021time, he2022timereplayer}. The visual comparison on HQF are shown in Figure.\ref{f6}. The comparison results tested on frames-plus-events datasets are shown in Table.\ref{tab5}. The proposed method outperforms the mainstream frame-only and frames-plus-events VFI methods on HighQualityFrames and HS-ERGB(close) datatsets. We achieve the second best results on HS-ERGB(far) in terms of SSIM. The two event-based VFI methods Time Lens and TimeReplayer are supposed to achieve the highest PSNR on HS-ERGB(far). However, we fail to get the claimed result for the former method, while the source of the later remains unavailable. 
	\begin{table*}[htbp]
		\scriptsize	\caption{Quantitative comparison tested on multiple event-based VFI benchmarks. Our proposed method and other baselines are tested on HighQualityFrames (HQF) and HS-ERGB datasets containing real-world events. The best PSNR(dB)\textbackslash SSIM result is \textcolor{red}{boldfaced}, and the second best is  \textcolor{blue}{\underline{underlined}}. $\ddagger$ indicates that we quote the results in corresponding papers. 
		}
		\label{tab5}
		\centering
		\setlength{\tabcolsep}{1.8pt}
		\renewcommand{\arraystretch}{1.2} 
		\begin{tabular}{lcccccccccccccc}
			
			\hline
			\toprule
			\multirow{3}*{Method}     &\multirow{3}*{Frame} &\multirow{3}*{Event} &\multicolumn{4}{c}{HQF \cite{stoffregen2020reducing}} &\multicolumn{4}{c}{HS-ERGB(far) \cite{tulyakov2021time}} &\multicolumn{4}{c}{HS-ERGB(close) \cite{tulyakov2021time}} \\
			\cline{4-15}
			\multicolumn{3}{c}{} &\multicolumn{2}{c}{1 frame skip} &\multicolumn{2}{c}{3 frames skip} &\multicolumn{2}{c}{5 frames skip} &\multicolumn{2}{c}{7 frames skip} &\multicolumn{2}{c}{5 frames skip} &\multicolumn{2}{c}{7 frames skip} \\
			\cline{4-15}
			\multicolumn{3}{c}{} & PSNR$\uparrow$ & SSIM$\uparrow$ &PSNR$\uparrow$ &SSIM$\uparrow$& PSNR$\uparrow$ & SSIM$\uparrow$ &PSNR$\uparrow$ &SSIM$\uparrow$ & PSNR$\uparrow$ & SSIM$\uparrow$ &PSNR$\uparrow$ &SSIM$\uparrow$ \\
			\midrule
			E2VID \cite{rebecq2019events}  &\XSolidBrush &\Checkmark   &10.02&0.302&10.36&0.301&11.46&0.508&11.43&0.505&8.89&0.396&9.03&0.399  \\
			DAIN \cite{bao2019depth}  &\Checkmark &\XSolidBrush&29.82  &0.875  &26.10 &0.782  & 27.92 &0.780 &27.13 &0.748 &29.03 &0.807&28.50 &0.801   \\
			RRIN \cite{li2020video} &\Checkmark &\XSolidBrush&29.76  &0.874 &26.11 &0.778 &25.62 &0.742 &24.14  &0.710  &28.69 &0.813 &27.46 &0.800 \\
			BMBC \cite{park2020bmbc}  &\Checkmark &\XSolidBrush &29.96  &0.875 &26.32 &0.781 &25.62 &0.742  &24.14  &0.710 &29.22 &0.820 &27.99 &0.808  \\
			AMBE \cite{park2021asymmetric}  &\Checkmark &\XSolidBrush&30.54 &0.891  &26.44 &0.798 &27.85&0.826&25.55&0.775&32.14&0.855&31.11&0.849\\
			VFIT-B \cite{shi2022video} &\Checkmark &\XSolidBrush &30.50 &0.882&-&-&-&-&-&-&-&-&-&-\\
			RIFE \cite{huang2022real}&\Checkmark &\XSolidBrush &32.26&0.889&28.08&0.796&29.46&0.845&27.18&0.797&32.98&0.865&31.77&0.855\\
			EMA-VFI\cite{zhang2023extracting}&\Checkmark &\XSolidBrush &31.42&0.885&27.76&0.802&29.70&0.857&27.49&0.807&\textcolor{blue}{\underline{33.49}}&\textcolor{blue}{\underline{0.869}}&\textcolor{blue}{\underline{32.38}}&\textcolor{blue}{\underline{0.859}}\\
			Time Lens \cite{tulyakov2021time}$\ddagger$  &\Checkmark &\Checkmark&\textcolor{blue}{\underline{32.49}}  &0.927 &\textcolor{blue}{\underline{30.57}}  &\textcolor{blue}{\underline{0.900}} &\textcolor{red}{\textbf{33.13}}  &\textcolor{red}{\textbf{0.877}} &\textcolor{red}{\textbf{32.31}} &\textcolor{red}{\textbf{0.869}}  &32.19 &0.839 &31.68 &0.835        \\
			TimeReplayer \cite{ he2022timereplayer}$\ddagger$&\Checkmark &\Checkmark&31.07&\textcolor{blue}{\underline{0.931}} &28.82 &0.866 &\textcolor{blue}{\underline{31.98}}&0.861&\textcolor{blue}{\underline{30.07}}&0.834 &31.21&0.818&29.83&0.816\\
			\textbf{Ours}    &\Checkmark &\Checkmark &\textcolor{red}{\textbf{32.74}}&\textcolor{red}{\textbf{0.934}}&\textcolor{red}{\textbf{31.40}}&\textcolor{red}{\textbf{0.913}}&30.65&\textcolor{blue}{\underline{0.874}} &28.94&\textcolor{blue}{\underline{0.841}}&\textcolor{red}{\textbf{33.58}}&\textcolor{red}{\textbf{0.871}}&\textcolor{red}{\textbf{32.75}}&\textcolor{red}{\textbf{0.864}}\\
			\bottomrule
			\hline
		\end{tabular}
	\end{table*}
	
	\subsection{Ablation Study}
	After step-by-step processing of each proposed module, the quality of output has been gradually improved. The results of each module are shown in Figure.\ref{01b}.
	
	\textbf{Effect of Gumbel Gating Module.} In order to verify the effectiveness of our proposed method for reducing computational consumption, we input all sub-graphs as dynamic regions into the residual optical flow estimation network, and the test results are shown in the third and fourth row of Table.\ref{tab3}. Compared with the method of considering all the sub-graphs as dynamic regions, the Tera-FLOPs and runtime of our proposed method tested on Vimeo90K-Triplet dataset greatly drops by 17\% and 10.6\% respectively, while the PSNR and SSIM merely decrease by 0.3dB and 0.001 individually. 
	
	\begin{table}[htbp]
		\small	
		\caption{Quantitative comparison of the proposed method of Gumbel gating module for calculating residual optical flow,  the method without  residual optical flow estimation module and the method of calculating residual optical flow of all the sub-regions. These methods are tested on Vimeo90K-Triplet validation set. Runtime is the total time to run the entire dataset. Tera-FLOPs indicates a trillion floating-point operations.}
		\label{tab3}
		\centering
		\setlength{\tabcolsep}{5pt}
		\begin{tabular}{lcccc}
			\hline
			\toprule
			Method &Runtime(s)$\downarrow$ &Tera-FLOPs$\downarrow$ &PSNR$\uparrow$ &SSIM$\uparrow$ \\
			\midrule
			Without refinement &263&0.145&37.19&0.967\\
			\midrule
			All regions process &442&0.253&39.40&0.977\\
			\textbf{Ours}    &395($\downarrow$10.6\%)  &0.210($\downarrow$17.0\%) &39.10 &0.976  \\
			\bottomrule
			\hline
		\end{tabular}
	\end{table}
	
	\textbf{Effect of Residual Optical Flow Estimation Module. }We input the warping-based frames $I_0^{warp}$, $I_1^{warp}$ generated on the rough optical flow estimation and the boundary frames $I_0$, $I_1$ into the final synthesis module, for verifying the effect of our proposed residual optical flow module on improving the final result. The experimental results are shown in the second row of Table.\ref{tab3}.
	
	\textbf{Effect of Cross-region Partition.} We set the size of the sliding window to be H/2$\times$W/2, which is consistent with our proposed method. In addition, the step size is set to H/2$\times$W/2 in the vertical and horizontal directions respectively, which will produce four regions without any cross-region parts. They are fed into our proposed residual optical flow estimation network for training. The test results on Vimeo90K are shown in Table.\ref{tab6}. Compared with the non-cross-region scheme, the proposed method with cross-region partition has a PSNR improvement of 1.68dB and a SSIM improvement of 0.008, which proves the effect of our method. As the cross-region part takes into account the correlation between domains, the optical flow estimation is smoother.
	\begin{table}[htbp]
		\small	
		\caption{Quantitative comparison of whether the region division includes cross-region partition on residual optical flow estimation module.}
		\label{tab6}
		\centering
		\setlength{\tabcolsep}{5pt}
		\begin{tabular}{lcc}
			\hline
			\toprule
			Region setting &PSNR$\uparrow$&SSIM$\uparrow$ \\
			\midrule
			Without Cross-region partition  &31.33&0.948\\
			Cross-region partition&33.01&0.956\\
			\bottomrule
			\hline
		\end{tabular}
	\end{table}

		\section{Conclusion and Discussion }
		We have proposed an event-and-frame-based VFI method for dynamically estimating optical flow and residual optical flow between adjacent frames, which maintains high-quality output while reducing computation time and overhead by 10\% and 17\% respectively. Tests on several large-scale VFI benchmark datasets show that our proposed method outperforms other state-of-the-art VFI methods in terms of PSNR and SSIM. \textbf{Limitations.} On account of the lack of photometric information for events, IDO-VFI performs as deficiently as other VFI methods in scenes with complex photometric changes. In the future, we will consider introducing a contrast maximization method and a photometric loss function to reconstruct sharp edges and luminosity in those challenging scenes. \textbf{Potential Negative Social Impacts.} The proposed method can be used in application scenarios such as modal analysis and monitoring. These applications may bring concerns such as public privacy and security issue to the society. Please use the VFI technology reasonably under the premise of complying with the laws and regulations of various countries.


\begin{thebibliography}{10}

\bibitem{posch2014retinomorphic}
Christoph Posch, Teresa Serrano-Gotarredona, Bernabe Linares-Barranco, and Tobi
  Delbruck.
\newblock Retinomorphic event-based vision sensors: bioinspired cameras with
  spiking output.
\newblock {\em Proc. IEEE}, 102(10):1470--1484, 2014.

\bibitem{sun2018pwc}
Deqing Sun, Xiaodong Yang, Ming-Yu Liu, and Jan Kautz.
\newblock Pwc-net: Cnns for optical flow using pyramid, warping, and cost
  volume.
\newblock In {\em Proceedings of the IEEE conference on computer vision and
  pattern recognition}, pages 8934--8943, 2018.

\bibitem{jiang2018super}
Huaizu Jiang, Deqing Sun, Varun Jampani, Ming-Hsuan Yang, Erik Learned-Miller,
  and Jan Kautz.
\newblock Super slomo: High quality estimation of multiple intermediate frames
  for video interpolation.
\newblock In {\em Proceedings of the IEEE conference on computer vision and
  pattern recognition}, pages 9000--9008, 2018.

\bibitem{bao2019depth}
Wenbo Bao, Wei-Sheng Lai, Chao Ma, Xiaoyun Zhang, Zhiyong Gao, and Ming-Hsuan
  Yang.
\newblock Depth-aware video frame interpolation.
\newblock In {\em Proceedings of the IEEE/CVF Conference on Computer Vision and
  Pattern Recognition}, pages 3703--3712, 2019.

\bibitem{park2020bmbc}
Junheum Park, Keunsoo Ko, Chul Lee, and Chang-Su Kim.
\newblock Bmbc: Bilateral motion estimation with bilateral cost volume for
  video interpolation.
\newblock In {\em European Conference on Computer Vision}, pages 109--125.
  Springer, 2020.

\bibitem{niklaus2020softmax}
Simon Niklaus and Feng Liu.
\newblock Softmax splatting for video frame interpolation.
\newblock In {\em Proceedings of the IEEE/CVF Conference on Computer Vision and
  Pattern Recognition}, pages 5437--5446, 2020.

\bibitem{sim2021xvfi}
Hyeonjun Sim, Jihyong Oh, and Munchurl Kim.
\newblock Xvfi: Extreme video frame interpolation.
\newblock In {\em Proceedings of the IEEE/CVF International Conference on
  Computer Vision}, pages 14489--14498, 2021.

\bibitem{yu2022deep}
Zhiyang Yu, Yu~Zhang, Xujie Xiang, Dongqing Zou, Xijun Chen, and Jimmy~S Ren.
\newblock Deep bayesian video frame interpolation.
\newblock In {\em Computer Vision--ECCV 2022: 17th European Conference, Tel
  Aviv, Israel, October 23--27, 2022, Proceedings, Part XV}, pages 144--160.
  Springer, 2022.

\bibitem{xu2019quadratic}
Xiangyu Xu, Li~Siyao, Wenxiu Sun, Qian Yin, and Ming-Hsuan Yang.
\newblock Quadratic video interpolation.
\newblock {\em Advances in Neural Information Processing Systems}, 32, 2019.

\bibitem{park2021asymmetric}
Junheum Park, Chul Lee, and Chang-Su Kim.
\newblock Asymmetric bilateral motion estimation for video frame interpolation.
\newblock In {\em Proceedings of the IEEE/CVF International Conference on
  Computer Vision}, pages 14539--14548, 2021.

\bibitem{lee2020adacof}
Hyeongmin Lee, Taeoh Kim, Tae-young Chung, Daehyun Pak, Yuseok Ban, and
  Sangyoun Lee.
\newblock Adacof: Adaptive collaboration of flows for video frame
  interpolation.
\newblock In {\em Proceedings of the IEEE/CVF Conference on Computer Vision and
  Pattern Recognition}, pages 5316--5325, 2020.

\bibitem{choi2021motion}
Myungsub Choi, Suyoung Lee, Heewon Kim, and Kyoung~Mu Lee.
\newblock Motion-aware dynamic architecture for efficient frame interpolation.
\newblock In {\em Proceedings of the IEEE/CVF International Conference on
  Computer Vision}, pages 13839--13848, 2021.

\bibitem{shi2022video}
Zhihao Shi, Xiangyu Xu, Xiaohong Liu, Jun Chen, and Ming-Hsuan Yang.
\newblock Video frame interpolation transformer.
\newblock In {\em Proceedings of the IEEE/CVF Conference on Computer Vision and
  Pattern Recognition}, pages 17482--17491, 2022.

\bibitem{niklaus2023splatting}
Simon Niklaus, Ping Hu, and Jiawen Chen.
\newblock Splatting-based synthesis for video frame interpolation.
\newblock In {\em Proceedings of the IEEE/CVF Winter Conference on Applications
  of Computer Vision}, pages 713--723, 2023.

\bibitem{shangguan2022learning}
Wentao Shangguan, Yu~Sun, Weijie Gan, and Ulugbek~S Kamilov.
\newblock Learning cross-video neural representations for high-quality frame
  interpolation.
\newblock In {\em Computer Vision--ECCV 2022: 17th European Conference, Tel
  Aviv, Israel, October 23--27, 2022, Proceedings, Part XV}, pages 511--528.
  Springer, 2022.

\bibitem{oh2022demfi}
Jihyong Oh and Munchurl Kim.
\newblock Demfi: deep joint deblurring and multi-frame interpolation with
  flow-guided attentive correlation and recursive boosting.
\newblock In {\em Computer Vision--ECCV 2022: 17th European Conference, Tel
  Aviv, Israel, October 23--27, 2022, Proceedings, Part VII}, pages 198--215.
  Springer, 2022.

\bibitem{pan2019bringing}
Liyuan Pan, Cedric Scheerlinck, Xin Yu, Richard Hartley, Miaomiao Liu, and
  Yuchao Dai.
\newblock Bringing a blurry frame alive at high frame-rate with an event
  camera.
\newblock In {\em Proceedings of the IEEE/CVF Conference on Computer Vision and
  Pattern Recognition}, pages 6820--6829, 2019.

\bibitem{wang2020joint}
Zihao~W Wang, Peiqi Duan, Oliver Cossairt, Aggelos Katsaggelos, Tiejun Huang,
  and Boxin Shi.
\newblock Joint filtering of intensity images and neuromorphic events for
  high-resolution noise-robust imaging.
\newblock In {\em Proceedings of the IEEE/CVF Conference on Computer Vision and
  Pattern Recognition}, pages 1609--1619, 2020.

\bibitem{yu2021training}
Zhiyang Yu, Yu~Zhang, Deyuan Liu, Dongqing Zou, Xijun Chen, Yebin Liu, and
  Jimmy~S Ren.
\newblock Training weakly supervised video frame interpolation with events.
\newblock In {\em Proceedings of the IEEE/CVF International Conference on
  Computer Vision}, pages 14589--14598, 2021.

\bibitem{tulyakov2021time}
Stepan Tulyakov, Daniel Gehrig, Stamatios Georgoulis, Julius Erbach, Mathias
  Gehrig, Yuanyou Li, and Davide Scaramuzza.
\newblock Time lens: Event-based video frame interpolation.
\newblock In {\em Proceedings of the IEEE/CVF Conference on Computer Vision and
  Pattern Recognition}, pages 16155--16164, 2021.

\bibitem{tulyakov2022time}
Stepan Tulyakov, Alfredo Bochicchio, Daniel Gehrig, Stamatios Georgoulis,
  Yuanyou Li, and Davide Scaramuzza.
\newblock Time lens++: Event-based frame interpolation with parametric
  non-linear flow and multi-scale fusion.
\newblock In {\em Proceedings of the IEEE/CVF Conference on Computer Vision and
  Pattern Recognition}, pages 17755--17764, 2022.

\bibitem{he2022timereplayer}
Weihua He, Kaichao You, Zhendong Qiao, Xu~Jia, Ziyang Zhang, Wenhui Wang,
  Huchuan Lu, Yaoyuan Wang, and Jianxing Liao.
\newblock Timereplayer: Unlocking the potential of event cameras for video
  interpolation.
\newblock In {\em Proceedings of the IEEE/CVF Conference on Computer Vision and
  Pattern Recognition}, pages 17804--17813, 2022.

\bibitem{zhang2022unifying}
Xiang Zhang and Lei Yu.
\newblock Unifying motion deblurring and frame interpolation with events.
\newblock In {\em Proceedings of the IEEE/CVF Conference on Computer Vision and
  Pattern Recognition}, pages 17765--17774, 2022.

\bibitem{wu2022video}
Song Wu, Kaichao You, Weihua He, Chen Yang, Yang Tian, Yaoyuan Wang, Ziyang
  Zhang, and Jianxing Liao.
\newblock Video interpolation by event-driven anisotropic adjustment of optical
  flow.
\newblock In {\em Computer Vision--ECCV 2022: 17th European Conference, Tel
  Aviv, Israel, October 23--27, 2022, Proceedings, Part VII}, pages 267--283.
  Springer, 2022.

\bibitem{zihao2018unsupervised}
Alex Zihao~Zhu, Liangzhe Yuan, Kenneth Chaney, and Kostas Daniilidis.
\newblock Unsupervised event-based optical flow using motion compensation.
\newblock In {\em Proceedings of the European Conference on Computer Vision
  (ECCV) Workshops}, pages 0--0, 2018.

\bibitem{jang2017categorical}
Eric Jang, Shixiang Gu, and Ben Poole.
\newblock Categorical reparametrization with gumble-softmax.
\newblock In {\em International Conference on Learning Representations (ICLR
  2017)}. OpenReview. net, 2017.

\bibitem{xue2019video}
Tianfan Xue, Baian Chen, Jiajun Wu, Donglai Wei, and William~T Freeman.
\newblock Video enhancement with task-oriented flow.
\newblock {\em International Journal of Computer Vision}, 127(8):1106--1125,
  2019.

\bibitem{baker2011database}
Simon Baker, Daniel Scharstein, JP~Lewis, Stefan Roth, Michael~J Black, and
  Richard Szeliski.
\newblock A database and evaluation methodology for optical flow.
\newblock {\em International Journal of Computer Vision}, 92(1):1--31, 2011.

\bibitem{nah2017deep}
Seungjun Nah, Tae Hyun~Kim, and Kyoung Mu~Lee.
\newblock Deep multi-scale convolutional neural network for dynamic scene
  deblurring.
\newblock In {\em Proceedings of the IEEE/CVF Conference on Computer Vision and
  Pattern Recognition}, pages 3883--3891, 2017.

\bibitem{rebecq2019events}
Henri Rebecq, Ren{\'e} Ranftl, Vladlen Koltun, and Davide Scaramuzza.
\newblock Events-to-video: Bringing modern computer vision to event cameras.
\newblock In {\em Proceedings of the IEEE/CVF Conference on Computer Vision and
  Pattern Recognition}, pages 3857--3866, 2019.

\bibitem{li2020video}
Haopeng Li, Yuan Yuan, and Qi~Wang.
\newblock Video frame interpolation via residue refinement.
\newblock In {\em Proceedings of the International Conference on Acoustics,
  Speech and Signal Processing}, pages 2613--2617. IEEE, 2020.

\bibitem{huang2022real}
Zhewei Huang, Tianyuan Zhang, Wen Heng, Boxin Shi, and Shuchang Zhou.
\newblock Real-time intermediate flow estimation for video frame interpolation.
\newblock In {\em Computer Vision--ECCV 2022: 17th European Conference, Tel
  Aviv, Israel, October 23--27, 2022, Proceedings, Part XIV}, pages 624--642.
  Springer, 2022.

\bibitem{zhang2023extracting}
Guozhen Zhang, Yuhan Zhu, Haonan Wang, Youxin Chen, Gangshan Wu, and Limin
  Wang.
\newblock Extracting motion and appearance via inter-frame attention for
  efficient video frame interpolation.
\newblock In {\em Proceedings of the IEEE/CVF Conference on Computer Vision and
  Pattern Recognition}, 2023.

\bibitem{park2023biformer}
Junheum Park, Jintae Kim, and Chang-Su Kim.
\newblock Biformer: Learning bilateral motion estimation via bilateral
  transformer for 4k video frame interpolation.
\newblock In {\em Proceedings of the IEEE/CVF Conference on Computer Vision and
  Pattern Recognition}, 2023.

\bibitem{rebecq2018esim}
Henri Rebecq, Daniel Gehrig, and Davide Scaramuzza.
\newblock Esim: an open event camera simulator.
\newblock In {\em Conference on robot learning}, pages 969--982. PMLR, 2018.

\bibitem{kingma2014adam}
Diederik~P Kingma and Jimmy Ba.
\newblock Adam: A method for stochastic optimization.
\newblock {\em arXiv preprint arXiv:1412.6980}, 2014.

\bibitem{stoffregen2020reducing}
Timo Stoffregen, Cedric Scheerlinck, Davide Scaramuzza, Tom Drummond, Nick
  Barnes, Lindsay Kleeman, and Robert Mahony.
\newblock Reducing the sim-to-real gap for event cameras.
\newblock In {\em Proceedings of the European Conference on Computer Vision},
  pages 534--549. Springer, 2020.

\bibitem{wang2004image}
Zhou Wang, Alan~C Bovik, Hamid~R Sheikh, and Eero~P Simoncelli.
\newblock Image quality assessment: from error visibility to structural
  similarity.
\newblock {\em IEEE Transactions on Image Processing}, 13(4):600--612, 2004.

\end{thebibliography}

	\end{document}